\documentclass[10pt,twocolumn,letterpaper]{article}

\usepackage{iccv}
\usepackage{times}
\usepackage{epsfig}
\usepackage{graphicx}
\usepackage{amsmath}
\usepackage{amssymb}
\usepackage{caption}

\usepackage[pagebackref=true,breaklinks=true,letterpaper=true,colorlinks,bookmarks=false]{hyperref}

\iccvfinalcopy 

\def\iccvPaperID{5892} 

\ificcvfinal\fi
\begin{document}

\title{Graph-Based Object Classification for Neuromorphic Vision Sensing}

\author{Yin Bi, Aaron Chadha, Alhabib Abbas, Eirina Bourtsoulatze and Yiannis Andreopoulos\\
Department of Electronic \& Electrical Engineering\\
University College London, London, U.K.\\
{\tt\small \{yin.bi.16, aaron.chadha.14, alhabib.abbas.13, e.bourtsoulatze, i.andreopoulos\}@ucl.ac.uk}
}

\maketitle

\begin{abstract}
Neuromorphic vision sensing (NVS)\ devices represent visual information as sequences of asynchronous discrete events (a.k.a., ``spikes'') in response to  changes in scene reflectance.
Unlike conventional active pixel sensing (APS), NVS allows for significantly higher event sampling rates at substantially increased energy efficiency and robustness to illumination changes. However, object classification with NVS streams cannot leverage on state-of-the-art convolutional neural networks (CNNs), since NVS does not produce frame representations. To circumvent this mismatch between sensing and processing with CNNs, we propose a compact graph representation for NVS. We couple this with novel residual graph CNN architectures and  show that, when trained on spatio-temporal NVS data for object classification, such residual graph CNNs preserve the spatial and temporal coherence of spike events, while requiring less computation and memory. Finally, to address the absence of large real-world NVS datasets for complex recognition tasks,  we   present and make available a  100k dataset of NVS recordings of the American sign language letters, acquired with an iniLabs DAVIS240c device under  real-world conditions.

\end{abstract}

\section{Introduction}

\begin{figure}
\centering
\includegraphics[width=3.3in]{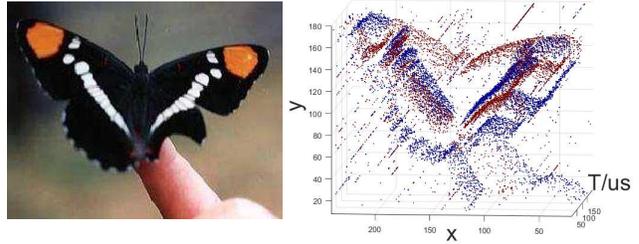}
\caption{Examples of objects captured by APS and neuromorphic vision sensors. Left: Conventional APS image. Right: Events stream from NVS\ sensor (Red:ON, Blue:OFF).}
\vspace{-0.2in}
\label{f:example}
\end{figure}

Object classification finds numerous applications in visual surveillance, human-machine interfaces, image retrieval and visual content analysis systems. Following the prevalence and advances of CMOS active pixel sensing (APS), deep convolutional neural networks (CNNs)  have already achieved good performance in APS-based object classification problems \cite{krizhevsky2012imagenet, he2016deep}. However, APS-based sensing is known to be cumbersome for machine learning systems because of limited frame rate,  high redundancy between frames, blurriness due to slow shutter adjustment under varying illumination, and  high power requirements \cite{tobi2016neu}. Inspired by the photoreceptor-bipolar-ganglion cell information flow in low-level mammalian vision, researchers have devised cameras based on neuromorphic vision sensing (NVS)   \cite{tobi2016neu,posch2011qvga, neftci2015neuromorphic}. NVS hardware outputs
a stream of asynchronous ON/OFF address events (a.k.a., ``spikes'')\ that indicate the changes in scene reflectance.  An example is  shown in Fig \ref{f:example}, where the NVS-based spike events correspond to a stream of coordinates and timestamps of reflectance events triggering ON or OFF in an asynchronous manner. This new principle significantly reduces the memory usage, power consumption and redundant information across time, while offering low latency and very high dynamic range.

However, it has been  recognized that current NVS-based object classification systems are inferior to APS-based counterparts because of the limited amount of work on NVS object classification and the lack of NVS data with reliable annotations to train and test with \cite{tobi2016neu, neftci2015neuromorphic, tan2015benchmarking}. In this work, we improve on these two issues by firstly proposing graph-based object classification method for NVS data. Previous approaches have either artificially grouped events into frame forms \cite{zhu2018ev, cohen2016event, cannici2018event, cannici2018attention} or derived complex feature descriptors \cite{sironi2018hats, clady2017motion, lagorce2017hots}, which do not always provide for good representations for complex tasks like object classification. Such approaches dilute the advantages of compactness and asynchronicity of NVS streams, and may be sensitive to the noise and change of camera motion or viewpoint orientation.  To the best of our knowledge, this is the first attempt to represent neuromorphic spike events as a graph, which allows us to use residual graph CNNs for end-to-end task training and reduces the computation of the proposed graph convolutional architecture to one-fifth of that of ResNet50 \cite{he2016deep}, while outperforming or matching the results of the state-of-the-art.

With respect to benchmarks, most neuromorphic datasets for object classification available to date are  generated
from emulators \cite{mueggler2017event, bi2017pix2nvs, garcia2016pydvs}, or recorded from  APS datasets via recordings of playback in standard  monitors \cite{serrano2015poker, li2017cifar10, hu2016dvs}. However, datasets acquired in this way  cannot capture scene reflectance changes as recorded by NVS\ devices in  real-world conditions. Therefore, creating real-world NVS\ datasets is important for the advancement of NVS-based computer vision. To this end, we create and make available a dataset of NVS recordings of 24 letters (A-Y, excluding J) from the American sign language. Our dataset provides more than 100K samples, and to our best  knowledge, this is the largest labeled NVS dataset acquired under realistic conditions.  

 We summarize our contributions as follows:

\begin{enumerate}
\item We propose a graph-based representation  for neuromorphic spike events, allowing for fast end-to-end task training and inference. 
\item We introduce \textit{residual graph CNNs} (RG-CNNs) for  NVS-based object classification. Our results show that they  require less computation and memory 
in comparison to conventional CNNs, while achieving
 superior results to the state-of-the-art in various datasets. 
\item We source one of the largest and most challenging neuromorphic vision datasets, acquired
under real-world conditions, and make it  available to the
research community. 
\end{enumerate}

In Section \ref{sec:related_work} we review related work. Section \ref{sec:methods} details our method for NVS-based object classification and is followed by the description of our proposed dataset in Section \ref{sec:datasets}. Experimental evaluation is presented in Section \ref{sec:experiments} and Section \ref{sec:conclusion} concludes the paper.

\begin{figure*}
  \vspace{-0.25in}
  \centering
  \includegraphics[width=6.0in]{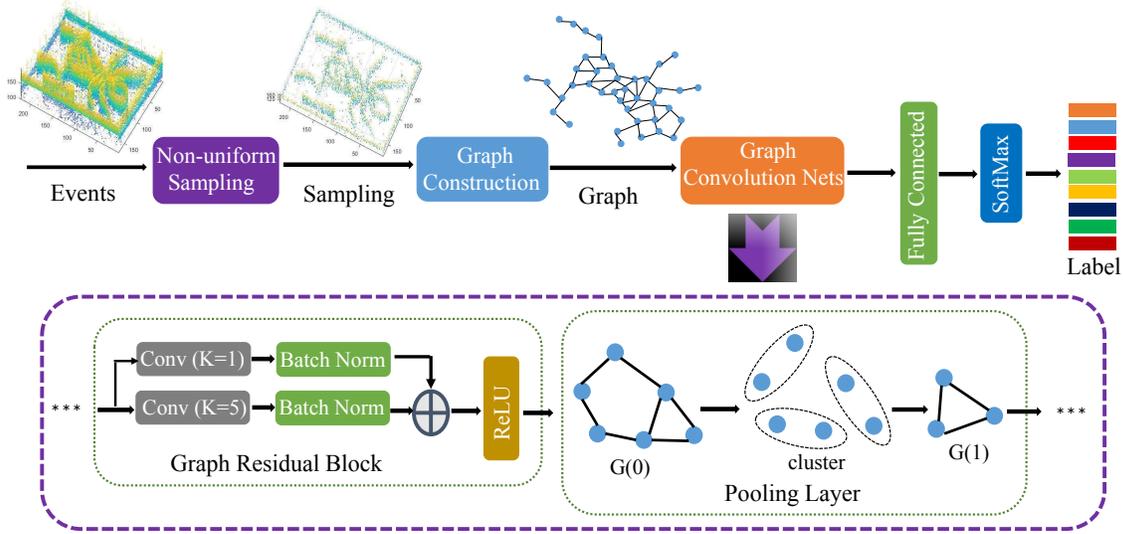}
  \vspace{-0.15in}
  \caption{Framework of graph-based object classification for neuromorphic vision sensing.}
	\vspace{-0.15in}
        \label{f:framework}
\end{figure*}

\section{Related Work}\label{sec:related_work}
We first review previous work on NVS-based object classification, followed by a review of recent developments in graph convolutional neural networks.
\subsection{Neuromorphic Object Classification }
Feature descriptors for object classification have been widely used by the neuromorphic vision community. Some of the most common descriptors are corner detectors and line/edge extraction \cite{clady2015asynchronous, vasco2016fast, mueggler2017fast, mueggler2017event}. While these efforts were promising early attempts for NVS-based object classification, their performance does not scale well when considering complex datasets. Inspired by their frame-based counterparts, optical flow methods have been proposed as feature descriptors for NVS \cite{clady2017motion, benosman2014event, barranco2014contour, barranco2015bio, brosch2015event, orchard2014bioinspired}. For a high-accuracy optical flow, these methods have very high computational requirements, which  diminishes their usability in real-time applications. In addition, due to the inherent discontinuity and irregular sampling of NVS data, deriving compact optical flow representations with enough descriptive power for accurate classification and tracking still remains a challenge \cite{clady2017motion, benosman2014event, brosch2015event, orchard2014bioinspired}. Lagorce proposed event-based spatio-temporal features called time-surfaces \cite{lagorce2017hots}. This is a time oriented approach to extract spatio-temporal features that are dependent on the direction and speed of motion of the objects. Inspired by time-surfaces, Sironi proposed a higher-order representation for local memory time surfaces that emphasizes the importance of using the information carried by past events to obtain a robust representation \cite{sironi2018hats}. A drawback of these methods is their sensitivity to noise and their strong dependencies on the type of motion of the objects in each scene. 

Another avenue for NVS-based object classification is via frame-based methods, i.e., converting the neuromorphic events to into synchronous frames of spike events, on which
conventional computer vision techniques can be applied. Zhu \cite{zhu2018ev} introduced a four-channel image form with the same resolution as the neuromorphic vision sensor: the first two channels encode the number of positive and negative events that have occurred at each pixel, while last two channels as the timestamp of the most recent positive and negative event. Inspired by the functioning of Spiking Neural Networks (SNNs) to maintain memory of past events, leaky frame integration has been used in recent work \cite{cohen2016event, cannici2018event, cannici2018attention}, where the corresponding position of the frame is incremented by a fixed amount when an event occurs at that event address. Peng \cite{peng2017bag} proposed bag-of-events (BOE) feature descriptors, which is a statistical learning method that firstly divides the event streams into multiple segments and then relies on joint probability distribution of the consecutive events to represent feature maps. However, these methods do not offer the compact and asynchronous nature of NVS, as the frame sizes that need to be processed are substantially larger than those of the original NVS\ streams. 

The final type of neuromorphic object classification is event-based methods. The most commonly used architecture is based on spiking neural networks (SNNs)  \cite{akopyan2015truenorth,diehl2015unsupervised, bichler2012extraction, lee2016training, neftci2014event}. While SNNs are theoretically capable of learning complex representations, they have still not achieved the performance of gradient-based methods because of lack of suitable training algorithms. Essentially, since the activation functions of spiking neurons are not differentiable, SNNs are not able to leverage on popular training methods such as backpropagation. To address this, researchers currently follow an intermediate step \cite{o2013real, diehl2015fast, perez2013mapping, stromatias2015scalable}: a neural network is trained off-line using continuous/rate-based neuronal models with state-of-the-art supervised training algorithms and then the trained architecture is mapped to an SNN. However, until now, despite their substantial implementation advantages at inference, the obtained solutions are complex to train and have typically achieved lower performance than gradient-based CNNs. Therefore, the proposed graph-based CNN approach for NVS can be seen as a way to bridge the compact, spike-based, asynchronous nature of NVS with the power of well-established learning methods for graph neural networks.

\subsection{Graph CNNs}
Generalizing neural networks to data with graph structures is an emerging topic in deep learning research. The principle of constructing CNNs on graph generally follows two streams: the spectral perspective \cite{kipf2016semi, levie2017cayleynets, defferrard2016convolutional, wang2018local, bruna2013spectral, sandryhaila2013discrete, shuman2012emerging} and the spatial perspective \cite{duvenaud2015convolutional, atwood2016diffusion, boscaini2016learning, fey2018splinecnn, masci2015geodesic, monti2017geometric}. Spectral convolution applies spectral filters on the spectral components of signals on vertices transformed by a graph Fourier transform, followed by spectral convolution.  Defferrard \cite{defferrard2016convolutional} provided efficient filtering algorithms by approximating spectral filters with Chebyshev polynomials that only aggregate local K-neighborhoods. This approach was further simplified by Kipf \cite{kipf2016semi}, who consider only the one-neighborhood for single-filter operation. Levie \cite{levie2017cayleynets} proposed a filter based on the Caley transform as an alternative for the Chebyshev approximation. As to spatial convolution, convolution filters are applied directly on the graph nodes and their neighbors. Several research groups have independently dealt with this problem. Duvenaud \cite{duvenaud2015convolutional} proposed to share the same weights among all edges by summing the signal over neighboring vertices followed by a weight matrix multiplication, while Atwood \cite{atwood2016diffusion} proposed to share weights based on the number of hops between two vertices. Finally, recent work \cite{monti2017geometric,fey2018splinecnn} makes use of the pseudo-coordinates of nodes as input to determine how the features are aggregated during locally aggregating feature values in a local patch. Spectral convolution operations require an identical graph as input, as well as complex numerical computations because they handle the whole graph simultaneously. Therefore, to remain computationally efficient,  our work follows the spirit of spatial graph convolution approaches and extends them to NVS data for object classification.

\section{Methodology}\label{sec:methods}

Our goal is to represent the  stream of spike events from neuromorphic vision sensors as a graph and perform convolution on the graph for object classification. Our model is visualized in Fig. \ref{f:framework}: a non-uniform sampling strategy is firstly used to obtain a small set of neuromorphic events for computationally and memory-efficient  processing; then sampling events are constructed into a radius neighborhood graph, which is processed by our proposed residual-graph CNNs for object classification. The details will be described in the following section.

\subsection{Non-uniform Sampling \&\ Graph Construction}

Given a NVS sensor with spatial address resolution of $H \times W$, we express a volume of events produced by an NVS camera as a tuple sequence: 
\begin{equation}
\{e_{i}\}_{N} = \{x_{i},y_{i},t_{i},p_{i}\}_{N}
\end{equation}
where $(x_{i},y_{i})\in \mathbb{R}^{H \times W}$ indicates the spatial address at which the spike event occured, $t_{i}$ is the timestamp indicating when the event was generated,  $p_{i} \in{{\{+1,-1\}}}$ is the event polarity (with +1, -1 signifying ON and OFF events respectively), and $N$ is the total number of the events.
To reduce the storage and computational cost, we use  non-uniform grid sampling  \cite{lee2001point} to sample a subset of $M$  representative events $\{e_{i}\}_{M}$ from $\{e_{i}\}_{N}$, where $M \ll N$. Effectively, one event is randomly selected from a space-time volume with the maximum number of events inside. If we consider $\mathbf{s}\{e_{i}\}_{i=1}^{k}$ to be such a  grid containing $k$ events, then only one event $e_{i}$ ($i \in [1,k]$) is randomly sampled in this space-time volume.
We then define the sampling events $\{e_{i}\}_{\{m\}}$ on a directed graph $\mathrm{G}^{}=\{\mathrm{\nu},\mathrm{\varepsilon},\mathrm{U}\}$, with $\mathrm{\nu}$ being the set of vertices, $\mathrm{\varepsilon}$ the set of the edges, and $\mathrm{U}$ containing pseudo-coordinates  that locally define the spatial relations between connected nodes.  The sampling events are independent and not linked, therefore, we regard each event $e_{i} : (x_{i},y_{i},t_{i},p_{i})$ as a node in the graph, such  that $\nu_{i}:(x_{i},y_{i},t_{i})$, with $\nu_{i} \in \mathrm{\nu}$. We define the connectivity of nodes in the graph based on the radius-neighborhood-graph strategy. Namely, neighboring nodes $\nu_{i}$ and $\nu_{j}$ are connected with an edge only if their weighted Euclidean distance $d_{i,j}$ is less than radius distance $\mathrm{R}$. For two spike events  $e_{i}$ and $e_{j}$, the Euclidean distance between them is defined as the weighted spatio-temporal distance: 
\begin{equation}
d_{i,j}=\sqrt{\alpha(|x_{i}-x_{j}|^{2}+|y_{i}-y_{j}|^{2})+\beta|t_{i}-t_{j}|^{2}}\leq \mathrm{R}
\end{equation}
where $\alpha$ and $\beta$ are weight parameters compensating for the difference in spatial and temporal grid resolution (timing accuracy is significantly higher in NVS cameras than spatial grid resolution). To limit the size of the graph, we constrain the maximum connectivity degree for each node by parameter $D_{\max}$. 

We subsequently define $u(i,j)$  for node  $i$, with connected node  $j$, as $u(i,j)=\left [ \left | x_{i}-x_{j} \right |, \left | y_{i}-y_{j} \right |\right ] \in \mathrm{U}$.After connecting all nodes of the graph $\mathrm{G}=\{\mathrm{\nu},\mathrm{\varepsilon}, \mathrm{U}\}$ via the above process, we consider the polarity of events as a signal that resides on the nodes of the graph $\mathrm{G}$. In other words, we define the input feature for each node  $i$, as $f^{(0)}(i)=p_i \in \{+1,-1\}$. 
\subsection{Graph Convolution}
Graph convolution generalizes the traditional convolutional operator to the graph domain. Similar to frame-based convolution,  graph convolution has two types \cite{bronstein2017geometric}: spectral and spatial. Spectral convolution  \cite{defferrard2016convolutional, wang2018local, bruna2013spectral, sandryhaila2013discrete, shuman2012emerging} defines the convolution operator by decomposing a graph in the spectral domain and then applying a spectral filter on the spectral components. However, this operation requires identical graph input and handles the whole graph simultaneously, so it is not suitable for  the variable and large graphs constructed from NVS. Spatial convolution  \cite{boscaini2016learning, fey2018splinecnn, masci2015geodesic, monti2017geometric} aggregates a new feature vector for each vertex using its neighborhood information weighted by a trainable kernel function. Because of this property, we consider  spatial convolution operation as a better choice when dealing with graphs from NVS. 

Similar to conventional frame-based convolution, spatial convolution operations on graphs are also an one-to-one mapping between kernel function and neighbors at relative positions w.r.t. the central node of the convolution. Let $i$ denote a node of the graph with feature $f(i)$, $\mathfrak{N}(i)$ denote the set of neighbors of node $i$ and  $g(u(i,j))$ denote the weight parameter constructed from the kernel function $g(.)$. The graph convolution operator $\otimes$ for this node can then be written in the following general form 
\begin{equation} \label{eq:graph_conv}
(f \otimes g)(i) = \frac{1}{\left | \mathfrak{N}(i) \right |}\sum_{j\in\mathfrak{N}(i)}f(j)\cdot g(u(i,j))
\end{equation}
where $\left | \mathfrak{N}(i) \right |$ is the cardinality of $\mathfrak{N}(i)$. We can generalize (\ref{eq:graph_conv}) to  multiple input features per node. Given the kernel function $\mathbf{g}=(g_{1},...,g_{l},...,g_{M_{in}})$ and input node feature vector $\mathbf{f}_l$, with $M_{in}$ feature maps indexed by $l$, the spatial convolution operation $\otimes$ for the node $i$ with $M_{in}$ feature maps is defined as:
\begin{equation}
(\mathbf{f}\otimes\mathbf{g})(i) = \frac{1}{\left | \mathfrak{N}(i) \right |} \sum_{\l=1}^{M_{in}}\sum_{j\in\mathfrak{N}(i)}f_{l}(j)\cdot g_{l}(u(i,j))
\label{eq:CONV}
\end{equation}

The kernel function $\mathbf{g}$ defines how to model the coordinates $\mathrm{U}$. The content of  $\mathrm{U}$ is used to determine how the features are aggregated and the content of $f_{l}(j)$ defines what is aggregated. Therefore, several spatial convolution operations \cite{boscaini2016learning, fey2018splinecnn, masci2015geodesic, monti2017geometric} on graphs were proposed by using different choice of kernel functions $\mathbf{g}$. Among them, SplineCNN \cite{fey2018splinecnn} achieves state-of-the-art results in several applications, so in our work we use the same kernel function as in SplineCNN. In this way, we leverage properties of B-spline bases to efficiently filter NVS\ graph inputs of arbitrary dimensionality. Let $((N_{1,i}^{m})_{1\leq  i\leq k_{1}},...,(N_{d,i}^{m})_{1\leq  i\leq k_{d}})$ denote $d$ open $\mathbf{B}$-spline bases of degree $m$ with $\mathbf{k}=(k_{1},...,k_{d})$ defining $d$-dimensional kernel size \cite{piegl2012nurbs}. Let $w_{p,l} \in \mathbf{W}$ denote a trainable parameter for each element $p$ from the Cartesian product $\mathfrak{P}= (N_{1,i}^{m})_{i}\times\cdot \cdot \cdot \times (N_{d,i}^{m})_{i}$ of the B-spline bases and each of the $M_{in}$ input feature maps indexed by $l$. Then the kernel function $g_{l}:[a_{1},b_{1}]\times\cdot \cdot \cdot \times[a_{d},b_{d}]\rightarrow \mathfrak{R}$ is defined as:
\begin{equation}
g_{l}(\mathbf{u})=\sum_{p \in P}w_{p,l} \cdot \prod _{i=1}^{d}N_{i,p_{i}}(u_{i})
\end{equation}
We denote a graph convolution layer as $\mathrm{Conv}(M_{\mathrm{in}},M_{\mathrm{out}})$, where $M_{\mathrm{in}}$ is the number of input feature maps and $M_{\mathrm{out}}$ is the number of output feature maps indexed by $l^{'}$. Then, a graph convolution layer with bias $b_{l}$, activated by activation function $\xi (t)$, can be written as:
\begin{eqnarray}
\mathrm{Conv}_{l^{'}}  &=&  \xi( \frac{1}{\left | \mathfrak{N}(i) \right |} \sum_{\l=1}^{M_{\mathrm{in}}}\sum_{j\in\mathfrak{N}(i)}f_{l}(j)\cdot \sum_{p \in P}w_{p,l}  \\ & \cdot &  \prod _{i=1}^{d}N_{i,p_{i}}(u_{i}) + b_{l^{'}}) \nonumber  
\end{eqnarray}
where $l^{'}=1,..,M_{\mathrm{out}}$, indicates the $l^{'}$th output feature map. Given a series of $C$ graph convolutional layers $(\mathrm{Conv}^{(c)})_{c\in[0,C]}$, the $c$-th layer has corresponding input feature map $\mathbf{f}^{(c)}$ over all nodes, with the input feature for node $i$ of the first layer  $\mathrm{Conv}^{(0)}$,  $f^{(0)}(i)=p_i\in{{\{+1,-1\}}}$.    

Finally, to accelerate deep network training, we use batch normalization \cite{ioffe2015batch} before the activation function in each graph convolutional layer. That is, the whole node feature $f_{l^{'}}$ over $l^{'}$ channel map is normalized individually via:
\begin{equation}
f^{'}_{l^{'}}= \frac{f_{l}-E(f_{l^{'}})}{\sqrt{\mathrm{Var}(f_{l^{'}})+\epsilon }}\cdot \gamma +\beta
\end{equation}
where $l^{'}=1,..,M_{\mathrm{out}}$,  $E(f_{l^{'}})$ and $\mathrm{Var}(f_{l^{'}})$ denote mean and variance of $f_{l^{'}}$ respectively, $\epsilon$ is used to ensure normalization does not overflow when the variance is near zero, and $\gamma$ and $\beta$ represent trainable parameters.

\subsection{Pooling Layer}
The utility of a pooling layer is to compact feature representations, in order to preserve important information while discarding irrelevant details \cite{yang2016exploit}. In  conventional APS-oriented CNNs, because of the uniform sampling grid (e.g.,  regular pixel array in images), pooling layers can be easily implemented by performing a max, average, or sum operation over neighbouring features. Similar to recent work in graph pooling \cite{simonovsky2017dynamic}, we apply pooling in order to obtain a coarser NVS graph. As shown in the pooling layer of the Fig. \ref{f:framework}, we first derive fixed-size clusters for graphs based on the node coordinates, then aggregate all nodes within one cluster, followed by the computation of new coordinates and features for the new nodes. 

Given a graph representation, let us denote the  spatial coordinates for node $i$ as $(x'_{i},y'_{i})\in \mathbb{R}^{H' \times W'} $ and resolution as  $H' \times W'$. We define the cluster size as $s_h \times s_w$, which corresponds to the downscaling factor in the pooling layer, leading to  $\left \lceil \frac{H'}{s_h} \right \rceil \times \left \lceil \frac{W'}{s_w} \right \rceil$  clusters. Given there are $\mathrm{num}$ nodes 
in one cluster, only one new node is generated on each cluster. For this new node, the coordinates 
are the average of coordinates of these $\mathrm{num}$ nodes, and the feature is the average or maximum of feature of these $\mathrm{num}$ nodes, according to whether a max pooling $(\mathrm{MaxP})$ or average pooling $(\mathrm{AvgP})$ strategy is used. 
Importantly, if there are connected nodes between two clusters, we assume the new generated nodes in these two clusters are connected with an edge. 

\subsection{Fully Connected Layer}
Given $M_{\mathrm{in}}$ feature maps $\mathbf{f}\longrightarrow\mathbb{R}^{P \times M_{\mathrm{in}}}$ from a graph with $P$ nodes, similar to CNNs, a fully connected layer in a graph convolutional network is a linear combination of weights linking all input features to outputs. Let us denote $f_{l}^{p}(x)$ as the feature in $l$th feature map of the $p$th node, then we can derive a fully connected layer for $q=1,...,Q$ as:
\begin{equation}
\label{eq:fout_q}f_{q}^{\mathrm{out}}(x)=\xi (\sum_{p=1}^{P}\sum_{l=1}^{M_{\mathrm{in}}}F_{P \times M_{in} \times Q}f_{l}^{p}(x)) \quad 
\end{equation}
where $Q$ is the number output channels indexed by $q$, $F$ is trainable weight with size $\mathrm{P \times M_{\mathrm{in}} \times Q}$, $\xi (t)$ is the non-linear activation function, e.g. ReLU: $\xi (t) = \max{(0,t)}$. For the remainder of the paper, we use  $\mathrm{FC}(Q)$ to indicate a fully connected layer with $Q$ output dimensions, comprising the results of \eqref{eq:fout_q}.

\subsection{Residual Graph CNNs}
Inspired by the idea of ResNet \cite{he2016deep}, we propose  residual graph CNNs in order  to resolve the well-known degradation problem inherent with increasing number of layers (depth) in graph CNNs \cite{li2018deeper}.   We apply residual connections for NVS-based object classification, as shown in the related block of Fig. \ref{f:framework}. Consider the plain (non-residual) baseline is a graph convolutional layer with the kernel size of 5 in each dimension, followed by a batch normalization \cite{ioffe2015batch} that accelerates the convergence of the learning process. We  consider a ``shortcut'' connection as a graph convolution layer with kernel size of 1 in each dimension, which matches the dimension of the output future maps, and is also followed by batch normalization. Then we perform element-wise addition of the node feature between shortcut and the baseline, with ReLU activation function.
We denote the resulting  graph residual block as  $\mathrm{Res_{g}(c_{in},c_{out})}$, with $c_{\mathrm{in}}$ input feature maps and $c_{\mathrm{out}}$ output feature maps.

We follow the common architectural pattern for feed-forward networks of interlaced convolution layers and pooling layers topped by fully-connected layers. For an input graph, a single convolutional layer is firstly applied, followed by batch normalization, and max pooling. This is then followed by $L$ graph residual blocks, each followed by a max pooling layer. Finally, two fully connected layers map the features to classes. For example, for $L=2$, we have the following architecture: $\mathrm{Conv}$ $\longrightarrow$ $\mathrm{MaxP}$ $\longrightarrow$ $\mathrm{Res_{g}}$ $\longrightarrow$ $\mathrm{MaxP}$ $\longrightarrow$ $\mathrm{Res_{g}}$ $\longrightarrow$ $\mathrm{MaxP}$ $\longrightarrow$ $\mathrm{FC}$ $\longrightarrow$ $\mathrm{FC}$.

\begin{figure}
\vspace{-0.05in}
  \centering
  \includegraphics[width=3.in]{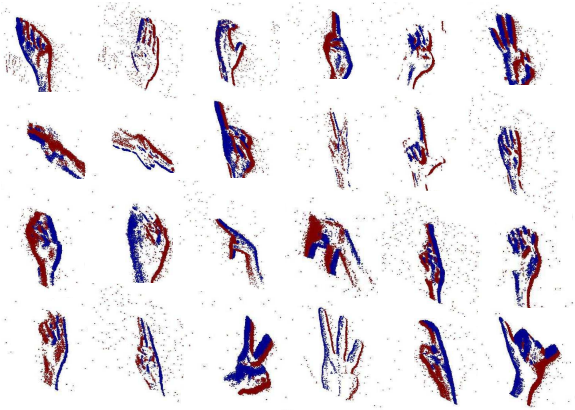}
  \vspace{-0.05in}
  \caption{Examples of the ASL-DVS dataset (the visualizations correspond to letters A-Y, excluding J, since letters  J and Z involve motion rather than static shape). Events are grouped to image form for visualization (Red/Blue: ON/OFF events).}
        \label{f:ASL}
\end{figure}

\begin{figure}
\vspace{-0.15in}
  \centering
  \includegraphics[width=3.in]{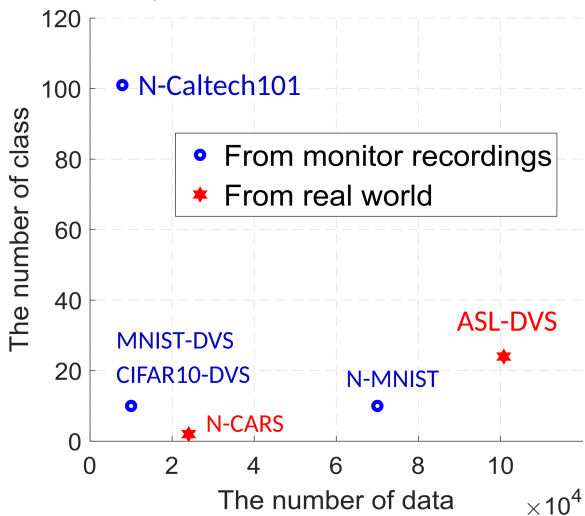}
  \vspace{-0.1in}
  \caption{Comparison of proposed NVS dataset w.r.t. the number of class and the number of total size. }
                \label{f:Datasets}
\vspace{-0.15in}
\end{figure}

\section{Datasets} \label{sec:datasets}
In this section, we first describe the existing NVS object classification datasets  and then we introduce our dataset that provides for an enlarged pool of NVS\ training and testing examples for handshape classification. 

\subsection{Existing Neuromorphic Datasets}
Many neuromorphic datasets for object classification are converted from standard frame-based datasets, such as N-MNIST \cite{orchard2015converting}, N-Caltech101 \cite{orchard2015converting}, MNIST-DVS \cite{serrano2015poker} and CIFAR10-DVS \cite{li2017cifar10}. N-MNIST and N-Caltech101 were acquired by an ATIS sensor \cite{posch2011qvga} moving in front of an LCD monitor while the monitor is displaying each sample image. Similarly, MNIST-DVS and CIFAR10-DVS datasets were created by displaying a moving image on a monitor and recording with a fixed DAVIS sensor \cite{lichtsteiner2008128}. Emulator software has also been proposed in order  to generate neuromorphic events from pixel-domain video formats using the change of pixel intensities of successively rendered images \cite{mueggler2017event, bi2017pix2nvs, garcia2016pydvs}.  While useful for early-stage evaluation, these datasets  cannot capture the real dynamics of an NVS\ device due to the limited frame rate of the utilized  content, as well as the limitations and artificial noise imposed by the recording or emulation environment. To overcome these limitations, N-CARS dataset \cite{sironi2018hats} was created by directly recording objects in urban environments with an ATIS sensor. This  two-class real-world dataset comprises 12,336 car samples and 11,693 non-car samples (background) with 0.1 second length. Despite its size, given that it only corresponds to a binary classifier problem, N-CARS cannot represent the behaviour of object classification algorithms on more complex NVS-based tasks.

\subsection{ American Sign Language Dataset (ASL-DVS)}
We present a large 24-class dataset  of handshape recordings under realistic conditions. Its 24 classes correspond to 24 letters (A-Y, excluding J) from the American Sign Language (ASL), which we call ASL-DVS. Examples of recordings  are shown in Fig \ref{f:ASL}. The ASL-DVS was recorded with an iniLabs DAVIS240c NVS camera set up in an office environment with low environmental noise and constant illumination. For all recordings, the camera was at the same position and orientation to the persons carrying out the handshapes. Five subjects were asked to pose the different static handshapes relative to the camera in order to introduce natural variance into the dataset. For each letter, we collected 4,200 samples (total of 100,800 samples) and each sample lasts for approximately 100 milliseconds. As is evident from Fig. \ref{f:ASL}, our ASL-DVS dataset presents a challenging task for event-based
classifiers, due to the subtle differences between the finger positioning of certain letters, such as N and O (first
two letters in row 3). Fig. \ref{f:Datasets} shows a comparison of existing NVS\ datasets w.r.t. the number of classes and total size. Within the landscape of existing datasets, our ASL-DVS is a comparably complex dataset  with the largest number of labelled examples. We  therefore hope that this will make it a useful resource for researchers to build comprehensive model for NVS-based object recognition, especially given the fact that it comprises real-world recordings. ASL-DVS and related code can be found at this link: \href{https://github.com/PIX2NVS/NVS2Graph}{https://github.com/PIX2NVS/NVS2Graph}.

\begin{table*}[ht]
\centering
\caption{Top-1 acccuracy of our CNNs w.r.t. the state of the art \& other graph convolution networks. }
\vspace{-0.1in}
\label{t:toART}
\begin{tabular}{cccccccc}
\hline
Model & N-MNIST & MNIST-DVS & N-Caltech101 & CIFAR10-DVS & N-CARS & ASL-DVS\\
\hline
H-First \cite{orchard2015hfirst} & 0.712 & 0.595 & 0.054 & 0.077 & 0.561 &-\\
HOTS \cite{lagorce2017hots} & 0.808 & 0.803 & 0.210 & 0.271 &0.624 &-\\
Gabor-SNN \cite{lee2016training,neil2016phased} & 0.837 & 0.824 & 0.196 & 0.245 & 0.789 &-\\
HATS \cite{sironi2018hats} & \textbf{0.991} & 0.984 & 0.642 & 0.524 & 0.902 &-\\
\hline
GIN \cite{xu2018powerful} & 0.754 & 0.719 & 0.476 & 0.423 & 0.846 & 0.514\\
ChebConv \cite{defferrard2016convolutional} & 0.949 & 0.935 & 0.524 & 0.452 & 0.855 & 0.317\\
GCN \cite{kipf2016semi} & 0.781 & 0.737 & 0.530 & 0.418 & 0.827 & 0.811\\
MoNet \cite{monti2017geometric} & 0.965 & 0.976 & 0.571 & 0.476 & 0.854 & 0.867\\
\hline
G-CNNs (this work) & 0.985 & 0.974 & 0.630 & 0.515 & 0.902& 0.875\\
RG-CNNs (this work) & 0.990 & \textbf{0.986} & \textbf{0.657} & \textbf{0.540} & \textbf{0.914} &\textbf{0.901}\\
\hline             
\end{tabular}
\end{table*}

\section{Experiments}\label{sec:experiments}

\subsection{Comparison to the State-of-the-Art} \label{s:toARTs}
In our experiments, the datasets of Fig. \ref{f:Datasets}  are used to validate our algorithm. For the N-MNIST, MNIST-DVS and N-CARS datasets, we use the predefined training and testing splits, while for N-Caltech101, CIFAR10-DVS and ASL-DVS, we follow the experiment setup of Sironi  \cite{sironi2018hats}: 20\% of the data is randomly selected for testing and the remaining is used for training. For each sample, we randomly extract a single 30-millisecond time window of  events, as input to our object classification framework.  During the non-uniform sampling, the maximal number of events $k$ in each space-time volume is set to 8. When constructing graphs, the radius $R$ is 3, weighted parameters $\alpha$ and $\beta$ are set to 1 and $0.5 \times 10^{-5}$, respectively, and the maximal connectivity degree $D_{\max}$ for each node is 32.

As to the architecture of graph convolution networks, we choose two residual graph blocks for simple datasets N-MNIST and MNIST-DVS ($L=2$). The architecture of our network for these datasets is $\mathrm{Conv(1,32)}$$\longrightarrow$$\mathrm{MaxP(32)}$$\longrightarrow$$\mathrm{Res_{g}(32,64)}$ $\longrightarrow$$\mathrm{MaxP(64)}$$\longrightarrow$$\mathrm{Res_{g}(64,128)}$$\longrightarrow$$\mathrm{MaxP(128)}$$\longrightarrow$ $\mathrm{FC(128)}$$\longrightarrow$$\mathrm{FC}(Q)$, with $Q$ is the number of classes of each dataset, and the cluster size in each pooling layer is 2$\times$2, 4$\times$4 and 7$\times$7, respectively. For the remaining datasets, three residual graph blocks  ($L$=3) are used, and the utilized network architecture is $\mathrm{Conv(1,64)}$$\longrightarrow$$\mathrm{MaxP(64)}$$\longrightarrow$$\mathrm{Res_{g}(64,128)}$$\longrightarrow$ $\mathrm{MaxP(128)}$$\longrightarrow$$\mathrm{Res_{g}(128,256)}$$\longrightarrow$$\mathrm{MaxP(256)}$$\longrightarrow$ $\mathrm{Res_{g}(256,512)}$$\longrightarrow$$\mathrm{MaxP(512)}$$\longrightarrow$$\mathrm{FC(1024)}$$\longrightarrow$$\mathrm{FC}(Q)$. Since the datasets are recorded from different sensors, the spatial resolution of each sensor is different (i.e., DAVIS240c: 240$\times$180, DAVIS128 \& ATIS: 128$\times$128), leading to various maximum coordinates for the graph.  We therefore set the cluster size in pooling layers in two categories; \textit{(i)} N-Caltech101 and ASL-DVS: 4$\times$3, 16$\times$12, 30$\times$23 and 60$\times$45; \textit{(ii)}\  CIFAR10-DVS and N-CARS: 4$\times$4, 6$\times$6, 20$\times$20 and 32$\times$32. 
We also compare the proposed residual graph networks (RG-CNNs) with their corresponding plain graph networks (G-CNNs) that stacked the same number of graph convolutional and pooling layers. The degree of B-spline bases $m$ of all convolutions in this work is set to 1.

In order to reduce overfitting, we add dropout with probability 0.5 after the first fully connected layer and also perform data augmentation. In particular, we spatially scale node positions by a randomly sampled factor within $[0.95,1)$, perform mirroring (randomly flip node positions along 0 and 1 axis with 0.5 probability) and rotate node positions around a specific axis by a randomly sampled factor within $[0,10]$ in each dimension. Networks are trained with the Adam optimizer  for 150 epochs, with batch size of 64 and learning rate of 0.001. The learning rate is decayed by a factor of 0.1 after 60 and 110 epochs. 

We compare Top-1 classification accuracy obtained from our model with that from HOTS \cite{lagorce2017hots}, H-First \cite{orchard2015hfirst}, SNN \cite{lee2016training,neil2016phased} and HATS \cite{sironi2018hats}. For SNN, the results are previously published, while for HOTS, H-First and  HATS, we report  results from Sironi  \cite{sironi2018hats}, since we use the same training and testing methodology. The results are shown in Table \ref{t:toART}. On five out of the six  evaluated datasets,  our proposed RG-CNNs consistently outperform these methods and sets a new state-of-the-art, achieving near-perfect classification on smaller datasets, N-MNIST and MNIST-DVS. 

Table \ref{t:toART}   also includes the classification results stemming from other graph convolutional networks. The architectures of all control networks are the same as our plain graph networks (G-CNNs) in this section, with the only difference being the graph convolutional operation. Here we consider four other graph convolution operations: GCN \cite{kipf2016semi}, ChebConv \cite{defferrard2016convolutional}, MoNet \cite{monti2017geometric} and GIN \cite{xu2018powerful}. The training details and data augmentation methods are the same as illustrated before. The classification accuracy stemming from all networks of Table \ref{t:toART} indicates that our proposed RG-CNN and G-CNN outperform all  other graph convolutional networks.

\begin{table*}
\centering
\caption{Top-1 acccuracy of our graph CNNs with graph input w.r.t. CNNs with image form input.}
\vspace{-0.1in}
\label{t:toCNNs}
\begin{tabular}{ccccccc}
\hline
Model & N-MNIST & MNIST-DVS & N-Caltech101 & CIFAR10-DVS & N-CARS & ASL-DVS\\
\hline
VGG\_19 \cite{simonyan2014very} & 0.972 & 0.983 & 0.549 & 0.334 & 0.728 & 0.806\\
Inception\_V4 \cite{szegedy2017inception} & 0.973 & 0.985 &0.578 & 0.379 & 0.864 & 0.832\\
ResNet\_50 \cite{he2016deep} & 0.984 & 0.982 & 0.637 &  \textbf{0.558} & 0.903 & 0.886\\
\hline
G-CNNs (this work) & 0.985 & 0.974 & 0.630 & 0.515 & 0.902 & 0.875\\
RG-CNNNs (this work) & \textbf{0.990} & \textbf{0.986} & \textbf{0.657} & 0.540 & \textbf{0.914} & \textbf{0.901}\\
\hline             
\end{tabular}
\end{table*}

\begin{table}
\caption{Complexity (GFLOPs) and size (MB) of networks.}
\vspace{-0.1in}
\label{t:flops}
\centering
\begin{tabular}{ccc}
\hline
Model & GFLOPs & Size (MB) \\
\hline
VGG\_19 \cite{simonyan2014very} & 19.63 & 143.65\\
Inception\_V4 \cite{szegedy2017inception} & 12.25 & 42.62 \\
ResNet\_50 \cite{he2016deep} & 3.87 & 25.61\\
\hline
G-CNNs & 0.39 & 18.81\\
RG-CNNs & 0.79 & 19.46\\
\hline             
\end{tabular}
\vspace{-0.1in}
\end{table}

\subsection{Comparison to Deep CNNs}
In order to further validate our proposal, we compare our results with conventional deep convolutional networks trained on event-based frames. We train/evaluate on three well-established CNNs; namely, VGG\_19 \cite{simonyan2014very}, Inception\_V4 \cite{szegedy2017inception} and ResNet\_50 \cite{he2016deep}. Given that the format of the required input for these CNNs is frame-based, we group neuromorphic spike events to frame form over a random time segment of 30ms, similar to the grouping images of Zhu \cite{zhu2018ev}. The two-channel event images have the same resolution as the NVS sensor, with each channel encoding the number of positive and negative events respectively at each position. To avoid overfitting, we supplement the training with heavy data augmentation: we resize the input images such that the smaller side is 256 and keep the aspect ratio, then randomly crop, flip and normalize 224$\times$224 spatial samples of the resized frame. We train all CNNs from scratch using stochastic gradient descent with momentum set to 0.9 and $L_{2}$ regularization set to $0.1 \times 10^{-4}$, and the learning rate is initialized at $10^{-3}$ and decayed by a factor of 0.1 every 10k iterations.  

The Top-1 classification accuracy of all networks is reported in Table \ref{t:toCNNs}, with the implementation of our proposed G-CNNs and RG-CNNs being the same as in Section \ref{s:toARTs}. As to reference networks, despite performing comprehensive data augmentation and $L_{2}$ regularization to avoid overfitting, the results acquired from conventional CNNs are still below the-state-of-the-art since event images contain far less information (see Fig. \ref{f:example}). However, the accuracy of our proposals surpasses that of conventional frame-based deep CNNs on nearly all datasets.

We now turn our attention to the complexity of our proposals and compare the number of floating-point operations (FLOPs) and the number of parameters of each model. In conventional CNNs, we compute FLOPs for convolution layers as \cite{molchanov2016pruning}:
\begin{equation}
\mathrm{FLOPs} = 2HW(C_{\mathrm{in}}K^{2}+1)C_{\mathrm{out}}
\end{equation}
where $H$, $W$ and $C_{\mathrm{in}}$ are height, width and the number of channels of the input feature map, $K$ is the kernel size, and $C_{\mathrm{out}}$ is the number of output channels. For graph convolution layers, FLOPs stem  from 3 parts  \cite{fey2018splinecnn}; \textit{(i)}\ for computation of B-spline bases, there are $N_{\mathrm{edge}}(m+1)^{d}$ threads each performing $7d$ FLOPs (4 additions and 3 multiplications), where $N_{\mathrm{edge}}$ is the number of edges, $m$ the B-spline basis degree and $d$ the dimension of graph coordinates; \textit{(ii)} for convolutional operations, the FLOPs count is $3N_{\mathrm{edge}}C_{\mathrm{in}}C_{\mathrm{out}}(m+1)^{d}$ , with factor 3 stemming from 1 addition and 2 multiplications in the inner loop of each kernel and  $C_{\mathrm{in}}$ and $C_{\mathrm{out}}$ is the number of input and output channels, respectively; \textit{(iii)} for scatter operations and the bias term, the FLOPs count is $(N_{\mathrm{edge}}+N_{\mathrm{node}})C_{\mathrm{out}}$, where $N_{\mathrm{node}}$ is the number of nodes. In total, we have 
\begin{eqnarray} \label{eq:GCN_FLOPs}
\mathrm{FLOPs} & = & N_{\mathrm{edge}}(m+1)^{d}(3C_{\mathrm{in}}C_{\mathrm{out}}+7d)
\nonumber \\ & + & (N_{\mathrm{edge}}+N_{\mathrm{node}})C_{\mathrm{out}}
\end{eqnarray}
For fully connected layers, in both conventional CNNs and GCNs, we compute FLOPs as \cite{molchanov2016pruning} $\mathrm{FLOPs} = (2I-1)O$,
 where $I$ is the input dimensionality and $O$ is the output dimensionality. With regards to the number of parameters, for each convolution layer in both CNNs and GCNs, it is $(C_{\mathrm{in}}K^{2}+1)C_{\mathrm{out}}$, while in fully connected layers, it is $(C_{\mathrm{in}}+1)C_{\mathrm{out}}$. As shown by \eqref{eq:GCN_FLOPs}, FLOPs of graph convolution  depend on the number of edges and nodes. Since the size of input graph varies per dataset, we opt to report representative results from N-Caltech101 in Table \ref{t:flops}. G-CNNS and RG-CNNs have a smaller number of weights and require the less computation compared to deep CNNs. The main reason is that the graph representation is compact, which in turn  reduces the amount of data needed to be processed. For N-Caltech101, the average number of nodes of each graph is  1,000, while grouping events into a  2-channel image makes the input size equal to 86,400.

\section{Conclusion}\label{sec:conclusion}
We propose and validate graph-based convolutional neural networks for  neuromorphic vision sensing based object classification. Our proposed plain and residual-graph based CNNs allow for condensed representations, which in turn allow for end-to-end task training and fast post-processing that matches the compact and non-uniform sampling of NVS hardware. Our results are shown to compete or outperform all other proposals on six datasets, and we propose and make available a new large-scale ASL dataset in order to motivate further progress in the field.

\textbf{Acknowledgments.} This work was funded by EPSRC, grants EP/R025290/1 and EP/P02243X/1, and European Union's Horizon 2020 research and innovation programme (Marie Sklodowska-Curie fellowship, grant agreement No. 750254).

\section{Supplementary Material}

In this supplementary material, we explore how performance and complexity is affected when varying the key parameters of our approach. Via the  ablation studies reported here, we justify the choice of parameters used for  our experiments in the paper. 

With regards to the parameters of the proposed graph CNNs, we experiment with the non-residual (i.e., plain) graph architecture (G-CNN) as a representative example, and explore the performance when varying the depth of graph convolution layer and the kernel size of graph convolution.  Concerning the graph construction, our studied parameters are the time interval under which we extract  events, the event sample size and the radius distance ($\mathrm{R}$) used to define the connectivity of the nodes.  All experiments reported in this supplementary note were conducted on the  N-Caltech101 dataset, since it has the highest number of classes among all datasets. Finally, training methods and data augmentation follow the description given in Section 5.1 of the paper. 

\subsection{Event Sample Size for Graph Construction}

The primary source of input compression is the non-uniform sampling of the events
prior to graph-construction, which is parameterized by $k$ in the paper.  We explore the effects of this input compression by varying $k$ and evaluating the accuracy to complexity (GFLOPs) tradeoff in Table \ref{t:sample}. No compression (i.e., $k$ =
1) gives accuracy/GFLOPs = 0.636/3.74, whereas increasing compression with $k$ = 12 gives accuracy/GFLOPs =
0.612/0.26 (i.e., 93\% complexity saving). This suggests that
the accuracy is relatively insensitive to compression up to
$k$ = 12 (with $k$ = 8 providing an optimal point) and it is
the graph CNN that provides for state-of-the-art accuracy.

\begin{table}[h]
\caption{Top-1 accuracy and complexity (GFLOPs) w.r.t. event sample size, parameterized by $k$.} \label{t:sample}
\centering
\begin{tabular}{cccc}
\hline
$k$ & Accuracy & GFLOPs \\
\hline
1 & 0.636 & 3.74  \\
8 & 0.630 & 0.39 \\
12 & 0.612 & 0.26 \\
 
\hline             
\end{tabular}
\vspace{-0.1in}
\end{table}

\subsection{Radius Distance }
When constructing graphs, the radius-neighborhood-graph strategy is used to define the connectivity of nodes. The radius distance ($\mathrm{R}$) is an important graph parameter: when the radius is large, the number of generated graph edges increases, i.e., the  graph becomes denser and  needs increased GFLOPs for the convolutional operations. On the other hand, if we set a small radius, the connectivity of nodes may decrease to the point that it does not represent the true spatio-temporal relations of events, which will harm the classification accuracy. In this ablation study, we varied the radius distance to $\mathrm{R} = \{1.5, 3, 4.5, 6\}$, to find the best distance with respect to  accuracy and complexity. The results are shown in Table \ref{t:distance}, where we demonstrate that  radius distance above 3 cannot improve the model performance while incurring  significantly increased complexity. Therefore, in our paper we set the radius distance to 3. Note that when radius distance changes from 4.5 to 6, the required computation increases only slightly because of the maximum connectivity degree $\mathrm{D_{max}}$ that is set to 32 to constrain the edge volume of graph. 

\begin{table}[h]
\caption{Top-1 accuracy and complexity (GFLOPs) w.r.t. radius distance} \label{t:distance}
\centering
\begin{tabular}{ccc}
\hline
Radius distance & Accuracy & GFLOPs \\
\hline
1.5 &  0.551 &  0.33 \\
3 & 0.630 & 0.39 \\
4.5 & 0.626  & 0.98 \\
6 & 0.624 & 1.19      \\
\hline             
\end{tabular}
\vspace{-0.1in}
\end{table}

\subsection{Time Interval of Events}
For each sample, events within a fixed time interval are randomly extracted to input to our object classification framework. In this study, we test under various time intervals, i.e., 10, 30, 50 and 70 milliseconds, to see their effect on the accuracy and computation. The results are shown in Table \ref{t:length}.  When extracting 30ms-events from one sample, the model achieves the highest accuracy, with modest increase in complexity over 10ms-events. Therefore, we opted for this setting in our paper.

\begin{table}[h]
\caption{Top-1 accuracy and complexity (GFLOPs) w.r.t. the length of extracted events} \label{t:length}
\vspace{-0.1in}
\centering
\begin{tabular}{ccc}
\hline
length (ms) & Accuracy & GFLOPs \\
\hline
10 & 0.528 & 0.31  \\
30 & 0.630 & 0.39 \\
50 & 0.613  & 0.92 \\
70 & 0.625 &  1.27\\
\hline             
\end{tabular}
\end{table}

\subsection{Depth of Graph Convolution Layers}
As to the architecture of graph convolution networks, experimental studies by Li \textit{et al.} \cite{li2018deeper} show that the model performance saturates or even drops when increasing  the number of layers beyond a certain point, since graph convolution essentially pushes representations of adjacent nodes
closer to each other. Therefore, the choice of depth of graph convolution layers ($\mathrm{D}$) affects the model performance as well as its size and its complexity. In the following experiment, we tested various depths from 2 to 6, each followed by a max pooling layer, and subsequently concluding the architecture with two fully connected layers. The number of output channels ($\mathrm{C_{out}}$) in each convolution layer and the cluster size ($[\mathrm{s_{h}},\mathrm{s_{w}}]$) in each pooling layers were as follows: \textit{(i)} $\mathrm{D}=2$: $\mathrm{C_{out}} = (128,2 56)$, $[{s_{h}},{s_{w}}]=(16 \times 12, 60 \times 45)$; \textit{(ii)} $\mathrm{D}=3$: $\mathrm{C_{out}} = (64, 128, 256)$, $[{s_{h}},{s_{w}}]=(8 \times 6, 16 \times 12, 60 \times 45)$; \textit{(iii)} $\mathrm{D}=4$: $\mathrm{C_{out}} = (64, 128, 256, 512)$, $[{s_{h}},{s_{w}}]=(4 \times 3, 16 \times 12, 30 \times 23, 60 \times 45)$; \textit{(iv)} $\mathrm{D}=5$: $\mathrm{C_{out}} = (64, 128, 256, 512, 512)$, $[{s_{h}},{s_{w}}]=(4 \times 3, 8 \times 6, 16 \times 12, 30 \times 23, 60 \times 45)$; \textit{(v)} $\mathrm{D}=6$: $\mathrm{C_{out}} = (64, 128, 256, 512, 512, 512)$, $[{s_{h}},{s_{w}}]=(2 \times 2, 4 \times 3, 8 \times 6, 16 \times 12, 30 \times 23, 60 \times 45)$. For all cases, the number of output channels of the two fully connected layers were 1024 and 101 respectively. The results are show in Table \ref{t:depth}: while the highest accuracy is obtained when the depth is $5$, complexity (GFLOPs) and size (MB) of the network is substantially increased in comparison to $\mathrm{D}=4$. Therefore, in our paper, we set the depth of graph convolution layer to $\mathrm{D}=4$. 

\begin{table}[h]
\caption{Top-1 accuracy, complexity (GFLOPs) and size (MB) of networks w.r.t. depth of convolution layer.} \label{t:depth}
\centering
\begin{tabular}{cccc}
\hline
Depth & Accuracy & GFLOPs & Size (MB) \\
\hline
2 & 0.514 & 0.11 & 5.53 \\
3 & 0.587 & 0.16 & 6.31 \\
4 & 0.630 & 0.39 & 18.81 \\
5 & 0.634 & 1.05 & 43.81 \\
6 & 0.615 & 2.99 & 68.81 \\
\hline             
\end{tabular}
\vspace{-0.1in}
\end{table}

\subsection{Kernel Size}
Kernel size determines how many neighboring nodes' features are aggregated into the output node. This comprises a tradeoff between model size and accuracy. Unlike conventional convolution, the number of FLOPs needed is independent of the kernel size. This is due to the local support property of the B-spline basis functions \cite{fey2018splinecnn}. Therefore we only report the accuracy and model size with respect to various kernel sizes.  In this comparison, the architecture is the same as the G-CNNs in Section 5.1, with the only difference being that the kernel size is increasing between 2 to 6. The results are shown in the Table \ref{t:kernel}. When kernel size is set as 3, 4, 5 and 6, the networks achieve the comparable accuracy, while the size of network  increases significantly when the kernel size increases. In our paper, we set kernel size in the graph convolution to 5, due to the slightly higher accuracy it achieves. It is important to note that, even with a kernel size of 5 that incurs a larger-size model in comparison to size of 3, our approach is still substantially less complex than conventional deep CNNs, as shown in Table 3  in our paper.

\begin{table}[h]
\caption{Top-1 accuracy and size (MB) of networks w.r.t. kernel size} \label{t:kernel}
\centering
\begin{tabular}{ccc}
\hline
Kernel size & Accuracy & Size (MB) \\
\hline
2 & 0.543 & 5.02  \\
3 & 0.626 & 8.30 \\
4 & 0.621  & 12.90 \\
5 & 0.630 & 18.81 \\
6 & 0.627 & 26.02 \\
\hline             
\end{tabular}
\vspace{-0.1in}
\end{table}

\subsection{Input Size for Deep CNNs}
We investigate how the input size controls the tradeoff between accuracy and complexity for conventional deep CNNs trained on event images. We follow the training protocol and event image construction described in Section 5.2 of the paper, but now downsize the event image inputs  to various resolutions prior to processing with the reference networks. The accuracy and complexity  (GFLOPs) is reported on N-Caltech101 in Table \ref{t:size}. ResNet-50 offers the
highest accuracy/GFLOPs tradeoff for conventional CNNs,
ranging from 0.637/3.87 to 0.517/0.28. However, our RG-CNN trained on graph inputs surpasses accuracy of ResNet-50 for all resolutions, whilst offering comparable complexity (0.79 GFLOPs).

\begin{table}
\caption{Accuracy/GFLOPs of networks w.r.t. input size on N-Caltech101, for conventional deep CNNs with event image inputs.}
\label{t:size}
\vspace{-0.1in}
\centering
\begin{tabular}{cccc}
\hline
Input Size & VGG\_19& Inception\_V4  & ResNet\_50 \\
\hline
 $224\times224$  & 0.549/19.63 & 0.578/9.24  &  0.637/3.87 \\
$112\times112$ & 0.457/4.93 & 0.4272/1.63 & 0.595/1.02 \\
$56\times56$ & 0.300/1.29 & 0.343/0.22 & 0.517/0.28 \\
\hline  
\hline
 G-CNNs & 0.630/0.39 & RG-CNNs & 0.657/0.79 \\
\hline 
\end{tabular}
\end{table}

{\small
\bibliographystyle{ieee_fullname}
\bibliography{egbib}
}

\

\end{document}